# Anomalous Frame Detection by Grouping Frame Similarities between Two Videos Computed by Vision-Language Model to Extract Expert Workers' Unique Actions

**Ryo Sakai[1*], Yongpeng Cao[2*] and Nobutaka Kimura[3]**

[1] Robotics Research Department, Research and Development Group, Hitachi, Ltd., Ibaraki, Japan
[2] School of Engineering, The University of Tokyo, Tokyo, Japan
[3] Next Research Department, Research and Development Group, Hitachi, Ltd., Kokubunji, Japan

Corresponding author: Ryo Sakai (e-mail: ryo.sakai.cx@hitachi.com).

**ABSTRACT** Maintenance of critical infrastructures, such as railways and power plants, is essential for operational safety and reliability. However, the declining number of skilled maintenance workers poses a serious challenge to sustaining these operations, highlighting the need to effectively transfer expert know-how to less experienced workers. Although traditional interview-based approaches have been used to elicit maintenance skills, they struggle to capture know-how that experts themselves may not consciously recognize. To address this gap, we proposed a method that detects anomalous frames of candidate actions including know-how by comparing a video of manual-based work with that of expert maintenance workers. In a simulated maintenance experiment involving a distribution board, our method targeted 11 types of actions not described in the manual and achieved a 66.9% extraction rate, marking a 50-percentage-point improvement over conventional techniques. These findings underscore the effectiveness of our approach in revealing hidden maintenance knowledge, thereby contributing to enhanced skill transfer and workforce development in critical infrastructure maintenance.

## I. INTRODUCTION

Maintenance inspection work is one of the critical tasks in the operation of infrastructure facilities such as railways and power plants. Since infrastructure facilities undergo repeated repairs and renovations over their long service lives, the maintenance tasks themselves have become increasingly complex. Meanwhile, there is a shortage of skilled maintenance personnel capable of performing these tasks adequately at each site [1, 2, 3]. One potential solution to this shortage is the training of new maintenance inspectors. Although new inspectors require time to re-experience the work performed by expert inspectors in order to acquire equivalent maintenance know-how, it is expected that understanding the expert's operational know-how could significantly shorten this period—as seen, for example, in cases where expert know-how in ship operations is used to accelerate personnel training [4]. In general, however, such expert know-how is often tacit and held only by the individual expert. If this know-how can be codified, it would be possible to educate maintenance procedures systematically, thereby efficiently increasing the pool of potential maintenance inspectors [5]. From these perspectives, there is a growing need to extract the work know-how from experts, convert it into explicit knowledge, and pass it on to less experienced workers.

One approach to codifying the work know-how possessed by experts is the interview technique employed by previous study [4, 6, 7, 8, 9, 10, 11]. In this approach, an interviewer prepares work-related questions and extracts the expert's know-how from their responses. This technique can only codify the work know-how that the interviewer can anticipate or that the expert is consciously aware of. On the other hand, it is challenging to obtain information regarding aspects of know-how that the interviewer does not anticipate or that the expert is unaware of. Moreover, actions that have become second nature to the expert—even if they are critical to the quality of work—may not be deliberately explained during the interview and thus remain unexpressed. Even these unconscious or undervalued elements, which may actually underpin the quality of the expert's work, should be codified if possible.

*Equal contribution.

One approach to codifying expert work know-how is to automatically extract candidate actions that reflect the expert's know-how from videos of their work. First, workers are instructed to perform tasks exactly as described in the manual and their actions are recorded on video. Next, experts perform the tasks as part of their daily work and their actions are similarly recorded. The differences obtained by comparing these two sets of videos represent the actions that deviate from the manual and are unique to experts, making them strong candidates for embodying the expert's know-how.

Related to this concept, technology for comparing videos to find differences such as added content or skipped portions has been studied for a long time [12, 13]. In these studies, deep learning-based methods are employed, and they can robustly detect notable differences even in the presence of variations in shooting angles and lighting conditions. In addition, with the recent advances in deep learning methods represented by Vision Language Models (VLMs), it is becoming possible to adjust the granularity of the events to be detected for each detection process. However, these methods require training on a sufficiently large dataset collected from the target work domain. While they can be applied to areas such as cooking where data can be collected from the Web, applying these methods to fields like infrastructure maintenance inspection is challenging because online information is scarce and data collection for each task is costly.

Therefore, this study aims to develop a method that detects anomalous frames within candidate actions, which may embody expert know-how. This is achieved by comparing work videos created based on manuals with videos of expert operations. In our approach, we assume that the process of codifying expert behavior involves detecting two types of deviations from the manual. The first is to detect actions that experts independently add when they feel it is necessary, and the second is to detect cases in which actions described in the manual are omitted because experts appropriately skip redundant tasks. To achieve this, this report proposes a method that creates a similarity map between video images using a pre-trained VLM and detects anomalous frames including the differential actions based on the characteristics of the similarity map. Our experiments, conducted on 11 types of actions not described in the manual mimicking real distribution board maintenance tasks, demonstrate that our proposed method achieves a 66.9% detection rate—approximately 50 percentage points higher than a conventional baseline. The contributions of this study are twofold:

i. We propose the method that leverages a pre-trained VLM to detect frames containing candidate know-how actions by comparing a video of expert with the video created based on manual, all without the need for additional training.
ii. In experiments of mimicking distribution board maintenance, our method achieved a higher detection rate than conventional baselines, thereby confirming its effectiveness.

The remainder of this paper is organized as follows. Section II reviews related work on video difference detection. Section III details problem definition. Section IV presents our proposed method, including the generation of the similarity map and deviation extraction procedures. Section V and VI explain the experimental design and results. Section VII presents the discussions of this study. Finally, Section VIII presents the conclusion of this study.

## II. RELATED WORK

Various methods have been proposed over the years for comparing two videos and extracting their differences. For example, there are methods that compute the similarity between two videos to create a similarity map and then detect anomalous frames from the videos based on this map [13, 14, 15, 16]. These methods generally consist of two major steps: constructing a similarity map and detecting differences using the similarity map. By employing deep learning–based techniques, they can robustly detect significant differences, even in the presence of noise such as variations in camera angles or lighting conditions, which should not be the focus.

In addition, as a method to extract differences specifically related to tasks within two videos, a technique using rapidly advancing large language models (LLM) has been proposed [12]. In this approach, the level of abstraction of the event to be detected as a difference is provided as a text query, allowing adjustments to be made for each detection process. This method directly detects differences by taking two videos as input.

However, these methods require the collection of large amounts of video data and training based on the collected data. While the cost of data collection can be reduced in areas where data is readily available on the web, fields such as maintenance inspection—where data is generally not publicly available and is scarce—face high training costs. Therefore, the ability to detect differences at minimal cost is crucial for practical applications.

In this study, to minimize the cost of real-world deployment, we propose a method for difference detection using a similarity map created by leveraging existing pre-trained VLM and captioning models.

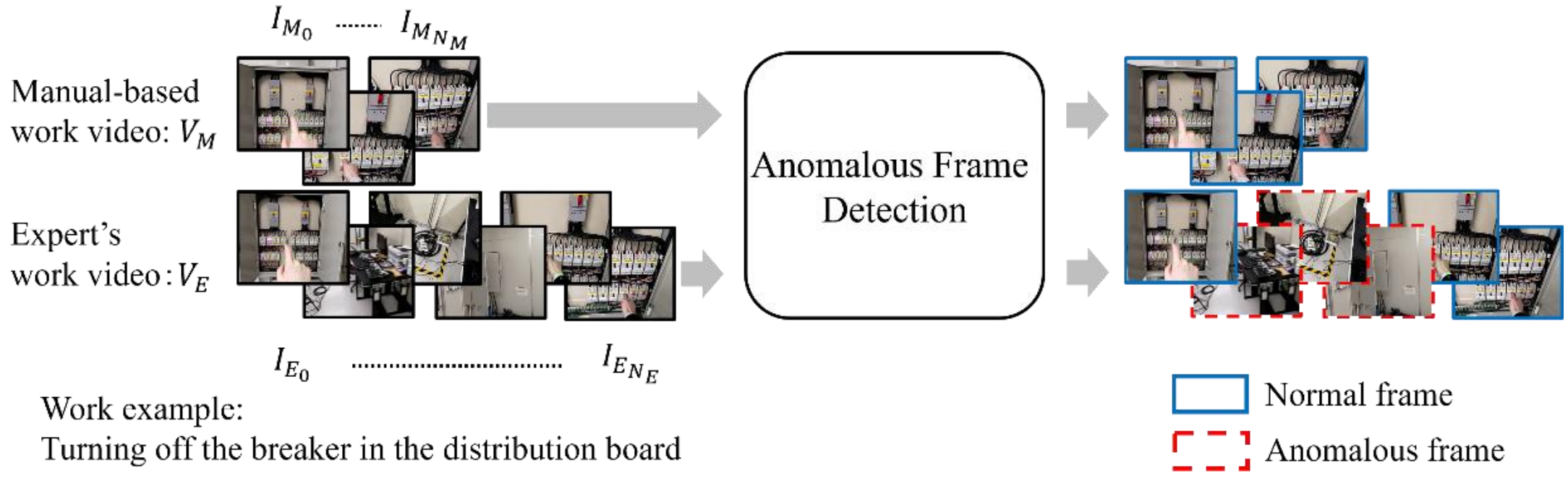

**FIGURE 1. Conceptual flow diagram illustrating the process of extracting candidate actions containing expert know-how from both the manual-based work video $V_M$ and the expert's work video $V_E$.**

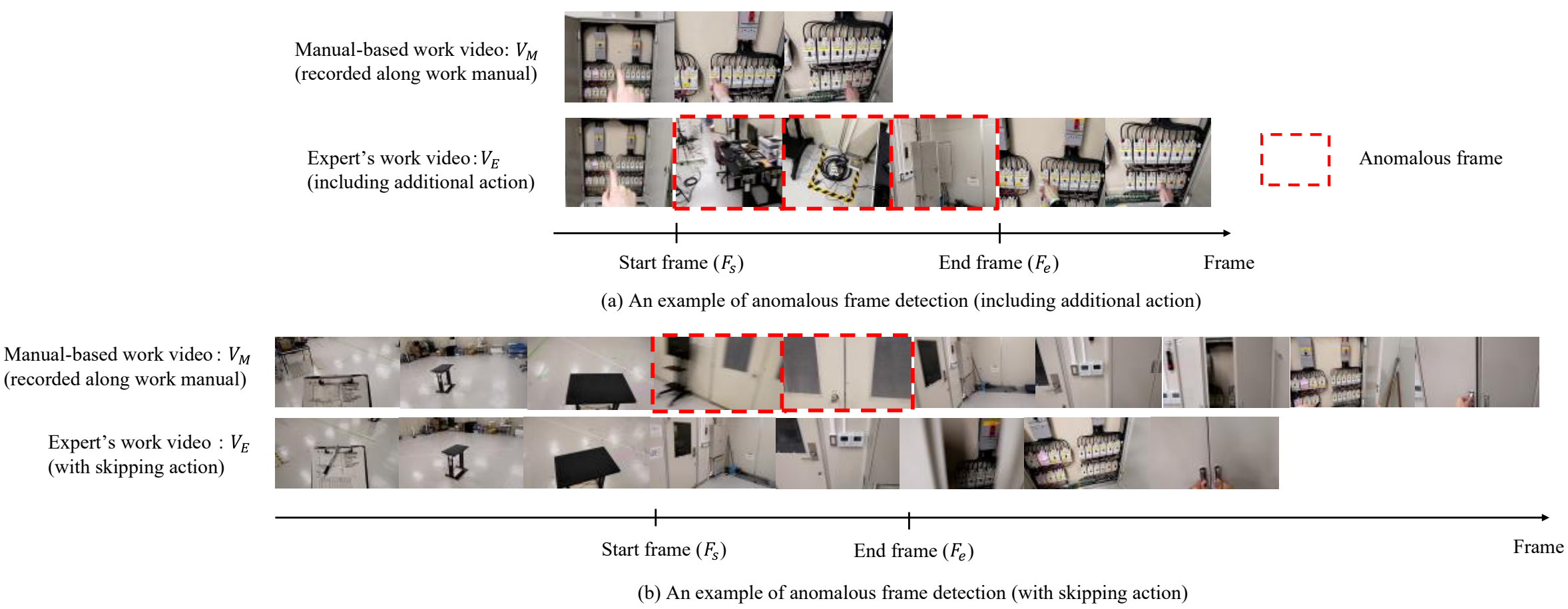

**FIGURE 2. Examples of the manual-based work video ($V_M$), the expert's work video ($V_E$) and the start frame ($F_s$) and end frames ($F_e$) to be detected. (a) An example of anomalous frame detection (including additional action). (b) An example of anomalous frame detection (with skipping action).**

A closely related area of interest in this study is video understanding. Recent works leverages pre-trained vision-language models (VLM) for video understanding [17]. For example, ActionCLIP [18] reformulates action recognition as video-text matching with prompt-based fine-tuning for zeroshot transfer, and subsequent works such as VideoLLM [19] harnesses decoder-only LLMs for unified reasoning across multiple video tasks. These works integrate pre-trained model knowledge for generalized action recognition, indicating the possibility of zero-shot action understanding.

However, these conventional methods require that actions be predefined and prepared as text or video clips so that they can be input as prompts or queries. Consequently, there is the additional burden of preparing these prompts or queries during practical use. Moreover, the way in which these actions are predefined can be ambiguous, potentially affecting estimation accuracy.

In contrast, the method developed in this study still utilizes VLMs as in these previous methods, but it extracts the differences between two videos simply by preparing the two videos.

## III. PROBLEM DEFINITION

The purpose of this study is to automatically extract candidate actions that include the expertise of skilled workers from both work videos created based on a manual and videos of expert operations. Figure 1 shows a conceptual flow diagram for extracting candidate actions that include expert know-how from the two videos mentioned above. Let the work video created based on the manual be represented as $V_M = \{I_{M_i} \mid 1 \le i \le N_M\}$ and the expert's operation video as $V_E = \{I_{E_j} \mid 1 \le j \le N_E\}$, where $I_{M_i}$ is the i-th frame image constituting $V_M$, and $I_{E_j}$ is the j-th frame image constituting $V_E$. Both $V_M$ and $V_E$ are first-person perspective videos of the worker.

In this study, in order to extract candidate actions that include expert know-how, we propose a method which takes $V_M$ and $V_E$ as inputs and detects the starting frame $F_s$ and the ending frame $F_e$ of the portions in $V_M$ and $V_E$ that capture these candidate actions. After detecting the starting frame $F_s$ and the ending frame $F_e$, the candidate actions that incorporate expert know-how can be extracted by detecting the images between these frames (the frames indicated by the red dashed line in Figure 1, hereinafter referred to as "anomalous frames"). The remaining images (the frames indicated by the blue solid line in Figure 1, hereinafter referred to as "normal frames") are considered to represent the actions performed by the expert according to the manual and can be regarded as actions that do not include the expert know-how.

FIGURE 3. Flowchart of the proposed method. The proposed method consists of the following two steps: (1) Generate a similarity map whose elements represent the similarities between each pair of frames from the manual-based work video $V_M$ and expert's work video $V_E$. In this similarity map, if a certain action appears only in one of the videos, it is expected to form a region with low similarity to all frames of the other video. (2) Identify regions of low similarity in the map and detect the corresponding frames as those containing actions that occur only in one of the videos.

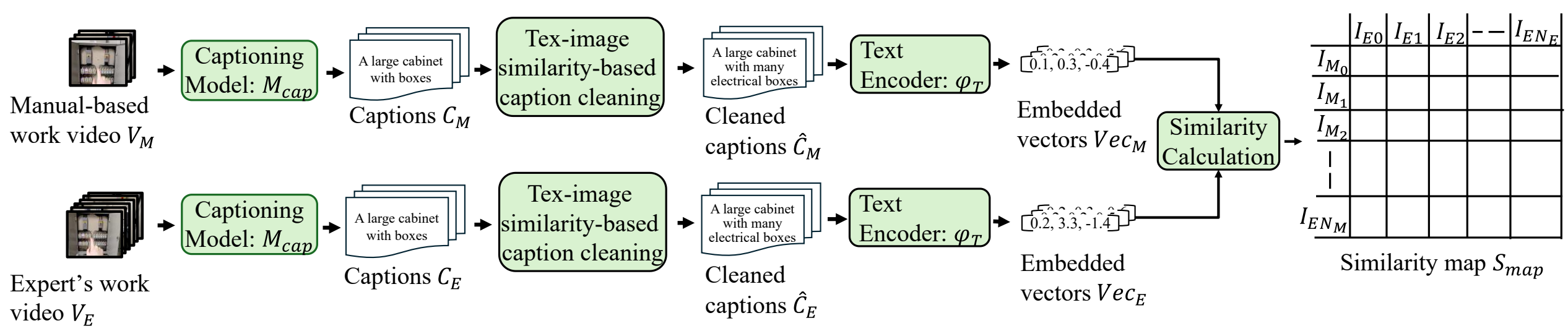

FIGURE 4. Workflow for creating similarity maps between videos $V_M$ and $V_E$. For each frame of the videos, the proposed method generates image captions $C_M$ and $C_E$ using a captioning model $M_{cap}$. To enhance the quality of the generated descriptions, the method vectorizes images and captions, compares their similarities to select the closest caption, and generates it as the cleaned caption. Then, the method vectorizes the cleaned captions $\hat{C}_M$ and $\hat{C}_E$ using a text encoder $\varphi_T$, calculates the similarity between each pair of vectors $Vec_M$ and $Vec_E$ and creates a similarity map $S_{map}$ between videos with calculated similarities as elements.

Figure 2 shows examples of $V_M$ and $V_E$, as well as the starting frame $F_s$ and the ending frame $F_e$, which are the targets for detection.

Figure 2(a) illustrates a case in which, as an example of work performed after business hours with the power to the distribution panel breaker being turned off, $V_E$ contains the action (A) that the expert has independently added because they felt it was necessary. $V_M$ is the work video created based on the manual for this operation, in which the operator carries out the breaker power-off as described in the manual. $V_E$ is the video recording of the expert's performed work. In this video, the action that the expert independently adds—namely, checking the outlet box connected to the distribution board before executing the breaker power-off—belongs to the anomalous frames of $V_E$ (the frames indicated by the red dashed line in Figure 2(a)). When such an added action exists, the objective is to detect the starting frame $F_s$ and the ending frame $F_e$ within the anomalous frames of $V_E$.

Figure 2(b) presents a case where, as an example of work performed before turning off the power to the distribution panel breaker due to a patrol inspection of the workroom, the action (B) described in the manual is not executed in $V_E$. Compared to $V_M$, the expert has skipped the action of confirming that the door is closed in $V_E$, and these frames belong to the anomalous frames of $V_M$ (the frames indicated by the red dashed line in Figure 2(b)). In the event that such a skipped action exists, the goal is to detect the starting frame $F_s$ and the ending frame $F_e$ within the anomalous frames of $V_M$.

As described above, experts generally follow the manual but may add actions not described in the manual or skip actions described in the manual. In this study, we assume that the type of deviation (i.e., an added action or a skipped action) is known for each video pair. This information can be obtained, for example, through expert annotation or prior task definition. Importantly, this assumption does not include any information about the temporal boundaries of the anomalous segment, which remain unknown and are the target of estimation.

## IV. PROPOSED METHOD

### A. Overview

Figure 3 shows an overview of the proposed method to detect the starting frame $F_s$ and the ending frame $F_e$, which indicate the segments in the work videos $V_M$ (created based on the manual) and $V_E$ (recording expert operations) that contain candidate actions with expert know-how. The proposed method detects $F_s$ and $F_e$ by comparing $V_M$ and $V_E$ and determining whether there exists a segment in which an action appears in only one of the videos (hereinafter referred to as an "anomalous action"). In order to realize this detection method, the proposed approach is composed of two steps.

In the first step, a similarity map is generated, where each element represents the similarity between the images constituting $V_M$ and $V_E$ (the green block in Figure 3). To calculate the similarity between images, a captioning model is employed to generate a description for each image, and the similarities between these generated descriptions are computed. It is expected that if similar actions are being performed in both videos, there will be regions of high similarity on the similarity map. On the other hand, if an anomalous action exists in only one of the videos, it is anticipated that there will be regions where the similarity is low for any image in the other video.

In the second step, regions with low similarity are identified from the similarity map to detect the segments that correspond to actions present in only one of the videos (the blue block in Figure 3). Specifically, filtering is applied to the similarity map to check whether the similarity meets certain conditions, and the beginning and end of the frames that satisfy these conditions are detected as the starting frame $F_s$ and the ending frame $F_e$, respectively.

Section IV-B describes the method for generating the similarity map using $V_M$ and $V_E$ as input. Next, Section IV-C explains the filtering method for determining the anomalous frames from the similarity map.

### B. Similarity Map Generation

Figure 4 shows the method for creating the similarity map between $V_M$ and $V_E$. First, captions are generated for the images, and these captions are converted into vectors using a text encoder. Then, by calculating the similarity using the vector representations of each image, the similarity map is created. Although it is possible to directly vectorize the images using an image encoder, it is expected that such a method would be significantly affected by changes in images that are less relevant to the work (e.g., differences in the shooting angle or lighting conditions). Therefore, in this study, captioning is performed on the images to robustly capture the work-related content within the images and to obtain vector representations of those captions.

Initially, using the captioning model $M_{cap}$, a caption corresponding to each image constituting the video is generated according to Equation (1).

$$C = \{c_i = M_{cap}(I_i) : 1 \le i \le N\} \quad (1)$$

Here, N is the number of images constituting the video, and $c_i$ is the caption corresponding to the i-th frame image. Further, $C$ is the set of captions $c_i$ for all images constituting the video. Caption generation is carried out for each image constituting $V_M$ and $V_E$, generating a set of captions $C_M$ for $V_M$ and a set of captions $C_E$ for $V_E$. The same captioning model is used for both $V_M$ and $V_E$.

Since $V_M$ and $V_E$ are videos recording a series of tasks, it is expected that if the images are temporally adjacent, the captions generated for the images will be similar in content. However, due to subtle differences between input images along the time axis, some captions generated for individual images might include content unrelated to the task, resulting in low-quality captions. These poor-quality captions may misrepresent the similarity of the work content recorded in the two videos when constructing the similarity map.

To address this problem and the potential random noisy output from the captioning model, the study creates a corrected set of captions, denoted as $\hat{C}$, by replacing poor-quality captions for a given image with more appropriate captions that are chosen from among the other captions. The corrected captions $\hat{C}$ are created based on the method described in previous work [20]. The outline of the method for generating $\hat{C}$ is explained using Equation (2):

$$\hat{c_i} = \underset{c \in C}{\operatorname{argmax}} < \varepsilon_I(I_i) \cdot \varepsilon_T(c) > \quad (2)$$

Here, $\varepsilon_I$ and $\varepsilon_T$ denote the image encoder and text encoder, respectively, Which are parts of the pre-trained multi-modal model. Since these encoders are trained to project images $I$ and texts $T$ into the same latent space $z$ within a multimodal foundation model, the latent vector $\varepsilon_I(I_i)$ and the latent vector $\varepsilon_T(c)$ can be regarded as vectors in the same space. Therefore, we can compute the similarity between image and text by multiplying these two vectors. In Equation (2), the symbol $<\cdot>$ denotes the measure of similarity. In this study, cosine similarity is used based on the approach in previous work [20]. By selecting the corrected caption $\hat{c_i}$ for each image, the sets of corrected captions $\hat{C}_M$ and $\hat{C}_E$ for $V_M$ and $V_E$ are produced.

Next, using the text encoder $\varphi_T$, we obtain the latent vector sets corresponding to the corrected caption sets $\hat{C}$, as shown in Equation (3). the latent vector sets $Vec_M$ and $Vec_E$ are produced for $V_M$ and $V_E$, respectively.

$$Vec = \{vec_i = \varphi_T(\hat{c_i}) : 1 \le i \le N\} \quad (3)$$

Then, cosine similarity is calculated between each pair of latent vectors, and a similarity map $S_{map}$ is created, with each element representing the computed similarity shown in Equation (4).

$$S_{map} = \{< vec_i \cdot vec_j > : 1 \le i \le N_M, 1 \le j \le N_E\} \quad (4)$$

$S_{map}$ is an $N_M \times N_E$ matrix, and the element in the i-th row and j-th column of $S_{map}$ corresponds to the similarity between the i-th frame image $I_{M_i}$ of $V_M$ and the j-th frame image $I_{E_j}$ of $V_E$.

### C. Anomalous Frame Filtering Based on Similarity Map

In the similarity map $S_{map}$, the similarity of frames in $V_E$ corresponding to the independently added action of experts (A) (as exemplified in Figure 2) or the similarity of frames in $V_M$ when the action described in the manual is not executed (B) becomes relatively small compared to the similarities of other frames. Therefore, the presence of anomalous actions can be inferred from locally low-similarity regions in the similarity map.

Based on this property, the proposed method detects anomalous frames by comparing the maximum similarity of each frame with a global reference value derived from the similarity map. Algorithm 1 shows the method for filtering anomalous frames based on the similarity map. To explain the filtering process clearly, we first describe the case where anomalous frames exist in $V_E$, i.e., when $V_E$ contains an additional action performed by the expert. The opposite case, where anomalous frames exist in $V_M$ due to an omitted action in $V_E$, is explained afterward. Figure 5 conceptually illustrates this filtering process, and the detailed computation is described below using Equation (5) and (6) in Algorithm 1.

First, using Equation (5), the average maximum similarity, $MaxAve$, of $S_{map}$ is computed. Each column of $S_{map}$ corresponds to an image constituting $V_E$. The similarity vector of the k-th column of $S_{map}$, denoted as $S_{map_k}$, has elements that indicate to what extent the k-th image $I_{E_k}$ of $V_E$ is similar to each image constituting $V_M$. When the expert is performing work in accordance with the manual, an image corresponding to $I_{E_k}$ should exist among the images constituting $V_M$, resulting in at least one high similarity value within $S_{map_k}$; ideally, the maximum similarity within $S_{map_k}$ would occur for the identical image. Therefore, in this study, the average of the maximum values of the similarity vectors along the column direction of $S_{map}$ is adopted as the criterion to judge whether an image $I_{E_k}$ is an anomalous frame, and this average is defined as the average maximum similarity $MaxAve$.

Next, Criterion (6) is used to determine whether $I_{E_k}$ is an anomalous frame. The threshold $\tau$ is a coefficient multiplied by the average maximum similarity $MaxAve$, and it is used to control the sensitivity of the proposed method to the differences between the two videos. Note that $\tau \geq 0$. When $\tau$ is high, even slight differences in similarity will lead to classifying a frame as anomalous, thereby extracting it as a candidate action that contains the expert know-how. If the maximum similarity in $S_{map_k}$ is lower than the product of $MaxAve$ and the threshold, then $I_{E_k}$ is regarded as an anomalous frame, and it is recorded into the anomalous frame cluster $F_{anomalous}$.

After applying the Criterion (6) for each column of $S_{map}$, the first frame in $F_{anomalous}$ is extracted by the function $GetFirstValue(F_{anomalous})$ and is considered to be the starting frame $F_S$. Additionally, the last frame in $F_{anomalous}$ is extracted by the function $GetEndValue(F_{anomalous})$ and is regarded as the ending frame $F_e$.

In the case where anomalous frames exist in $V_M$ – that is, when the action described in the manual is not executed (B) – $F_s$ and $F_e$ can be detected in a similar manner. Specifically, the same procedure is applied symmetrically by scanning direction in Algorithm 1 from column-wise to row-wise. In this way, by applying the process of Algorithm 1 in both the row and column directions of $S_{map}$, the frames corresponding to the candidate actions that include the expert know-how can be detected.

Figure 5 provides a conceptual illustration of the filtering process in Algorithm 1. Figure 5(a) corresponds to the case where $V_E$ contains the action (A) that the expert has independently added because they felt it was necessary, and the anomaly detection is performed column-wise. Figure 5(b) corresponds to the case where anomalous frames exist in $V_M$, i.e., when the action described in the manual is not executed, and the anomaly detection is performed row-wise. In both cases, the average maximum similarity is first computed by Equation (5), and then the anomaly criterion in Equation (6) is sequentially applied, as indicated by the arrows in the figure. The effectiveness of this filtering strategy is evaluated experimentally in Section V.

**Algorithm 1** Average filtering process for detecting anomalous frame

**Require:** $S_{map}$, $\tau$
**Ensure:** $F_S$, $F_e$

$$MaxAve \Leftarrow \frac{\sum_{l=0}^{N_E} \max(S_{map_l})}{N_E} \quad (5)$$

$F_{anomalous} = \phi$
**for** $k = 1$ to $N_E$ **do**
  **if** $\max(S_{map_k}) < \tau * MaxAve$ (6) **then**
    $F_{anomalous} = F_{anomalous} \cup k$
  **end if**
**end for**
$F_s \Leftarrow GetFirstValue(F_{anomalous})$
$F_e \Leftarrow GetEndValue(F_{anomalous})$

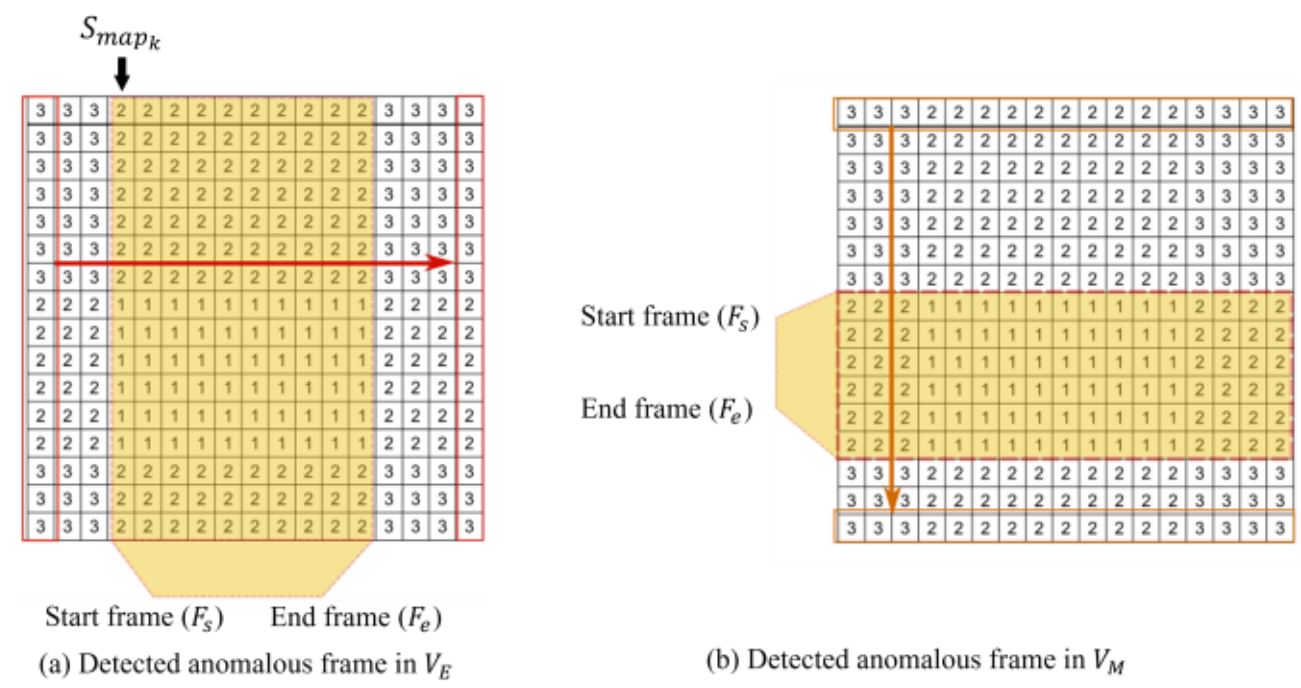

**FIGURE 5. Conceptual diagram of filtering to $S_{map}$. (a) An example of anomalous frames detecting in $V_E$. (b) An example of anomalous frames detecting in $V_M$.**

## V. EXPERIMENT

### A. Overview

In this study, using electrical panel maintenance operations as the subject, we created work videos based on the manual and videos of expert operations, and conducted an experiment to automatically extract candidate actions that include expert know-how.

First, the operator performed the work according to the manual, and a first-person perspective video $V_M$ was recorded during the process. Next, the operator performed the task with additional actions corresponding to expert operations, recording a video $V_E$ in which the behavior (A) that the expert independently felt was necessary was added. In addition, for an experiment intended to formalize the tacit knowledge of the expert by having the expert appropriately skip the redundant actions described in the manual, the operator recorded a video $V_E$ in which the action (B) specified in the manual was not performed.

In this experiment, by comparing the recorded $V_M$ and $V_E$, candidate actions incorporating the expert's know-how were extracted. To maintain the consistency, we ensured that each pair of videos contained at most one additional behavior (A) or one omitted behavior (B).

### B. Collected data

In this experiment, data were collected for two tasks in electrical panel maintenance operations: (task1) the operation to switch off the electrical panel breaker and (task2) the room inspection operation before switching off the panel. For (task1), as shown in the $V_M$ of Figure 2(a), the work defined as "checking the electrical panel breaker and switching it OFF" based on the manual was performed and recorded as $V_M$. In addition to these actions, in $V_E$ shown in Figure 2(a), the behavior of checking the connected equipment at the outlet box of the breaker before switching it off—corresponding to the expert's action (A)—was added, and $V_E$ was recorded. The recording of $V_M$ and $V_E$ for (task1) was carried out in two different buildings (Building 1 and Building 2): in Building 1, one operator's $V_M$ was collected along with three $V_E$'s; in Building 2, one set each of $V_M$ and $V_E$ was collected from two operators. Consequently, a total of 5 pairs of $V_M$ and $V_E$ were collected for (task1).

For (task2), as shown in the $V_M$ of Figure 2(a), the work defined as "sequentially checking the desktop, door, air conditioner power, and electrical panel in the workroom" based on the manual was performed and recorded as $V_M$. In addition to these actions, the following expert actions were added in $V_E$: (A-1) moving in front of the air conditioner to confirm that it is not operating, and (A-2) storing equipment if there is equipment on the desktop. For (task2), the recording of $V_M$ and $V_E$ was carried out in Building 1, in which one $V_M$ and one $V_E$ containing action (A-1) as well as one $V_E$ containing action (A-2) were collected from each of two operators. In addition, as a video recording of the expert's behavior of type (B), a $V_E$ was recorded in which the door was not checked. For this (B) recording, one $V_E$ was collected from each of the two operators. Consequently, a total of 6 pairs of $V_M$ and $V_E$ were collected for (task2), and when combined with (task1), a total of 11 pairs were collected. The operators wore a Microsoft Hololens2 integrated helmet [21] for data collection. The operators recorded data via a modified StreamRecorder app. from Hololens2ForCV library [22], modified to start and stop recording and to display holograms.

### C. Metrics

For each of the collected $V_M$ and $V_E$, manual annotations were made for the starting frame $F_{s_{gt}}$ and the ending frame $F_{e_{gt}}$ of each action in the video, thereby creating the ground truth data. In order to evaluate the automatic extraction of candidate actions that contain expert know-how, we employ the Intersection over Union(IoU) metric that calculates the degree to which the time interval defined by the automatically extracted starting frame $F_S$ and ending frame $F_E$ covers the interval in which the candidate action exists, as defined by the ground truth starting frame $F_{s_{gt}}$ and ending frame $F_{e_{gt}}$. This is shown in Equation (7).

$$Acc_{union} = \frac{(F_E - F_S + 1) \cap \left(F_{E_{gt}} - F_{S_{gt}} + 1\right)}{(F_E - F_S + 1) \cup \left(F_{E_{gt}} - F_{S_{gt}} + 1\right)} \quad (7)$$

In addition to Equation (5), the total number of frames that failed to be extracted was also computed. This number is the sum of the frames that should have been extracted as candidate actions but were missed and the frames that should not have been extracted as candidate actions but were erroneously extracted.

To verify the effectiveness of the proposed method, a comparison with a conventional method was conducted. As a conventional approach, we evaluated a method based on HLLM [20]—a technique that, similar to this report, aims to detect anomalous frames in the video by having a large language model (LLM) evaluate captions generated for the images. The captioning model in HLLM is BLIP-2 [23], and for cleaning the captions, the image encoder and text encoder used are the same as those in the paper, namely ImageBind [24]. In addition, an LLM (Llama3 [25]) is used to detect the starting frame $F_S$ and ending frame $F_E$ based on summarized captions. Unlike the conventional approach that focuses on detecting specific types of actions, our task does not target predefined action categories. Therefore, we implemented this method without the scoring system, instead directly prompting the LLM to detect the range of anomalous frames. The instruction prompts for the LLM were modified to match the problem setting in this report. Evaluations were conducted using Llama3.1-8B-Instruct [26] and Llama-3.2-3B-Instruct [27].

Furthermore, as an extension of HLLM [20], it is possible to generate captions to be evaluated by the LLM using a video-based captioning model rather than an image-based captioning model. Therefore, we also evaluated a method in which captions are generated using InternVideo2 [28]—which is capable of video-based caption generation—and the starting frame $F_S$ and ending frame $F_E$ are detected using Llama3.

To make the baseline comparison consistent with the objective of this study, we focused on methods that are directly applicable under the same zero-shot and training-free setting as the proposed framework. Specifically, the HLLM-based method represents a caption-based anomaly localization pipeline using LLM reasoning, while the InternVideo2-based variant serves as a stronger video-understanding baseline using video-level caption generation.

In the proposed method, the captioning model $M_{cap}$ used is, like in HLLM, BLIP-2 [23]. The text encoder $\varphi_T$ is ImageBind [24]. We intentionally adopted the same BLIP-2 and ImageBind configuration as the HLLM-based baseline so that the comparison focuses on the effect of the proposed similarity-map construction and filter-based localization, rather than differences in the encoder architecture. In Algorithm 1, we set the threshold $\tau$ to 0.6 based on the results of the sensitivity analysis described in Section V-D, which evaluates the estimation performance for different values of $\tau$.

### D. Sensitivity Analysis of Threshold $\tau$

To investigate the impact of the threshold $\tau$ in Equation (6) on the detection performance and to determine an appropriate value for our main comparative evaluations, we conducted a sensitivity analysis. The parameter $\tau$ controls the method's sensitivity to video differences; a higher $\tau$ detects even slight deviations, while a lower $\tau$ requires substantial differences to flag an anomaly. We evaluated the extraction performance by varying $\tau$ from 0.0 to 1.0 in increments of 0.1, assessing both the $Acc_{union}$ and the number of failed frames.

### E. Ablation Studies

To evaluate the effect of the caption correction step in Eq. (2), we conducted an ablation study comparing the use of raw captions (Eq. (1)) and cleaned captions (Eq. (2)). In Algorithm 1, we set the threshold $\tau$ to 0.6. The results show that the use of cleaned captions improves the union accuracy from 0.617 to 0.669 and reduces the number of failed frames from 231.09 to 213.45. These results suggest that the correction step effectively reduces noise and inconsistencies in the captions caused by minor visual variations. By selecting captions that are more consistent with the shared image–text embedding space, Eq. (2) produces a more stable similarity map, leading to more accurate detection of anomalous segments, particularly in boundary estimation.

## VI. RESULTS

Figure 6 shows an example of the similarity map $S_{map}$ generated by the proposed method from $V_M$ and $V_E$ for (task1). As illustrated in Figure 6, the action of switching off the electrical panel breaker appears in both $V_M$ and $V_E$, and the corresponding region in the similarity map $S_{map}$ exhibits a high similarity. In this way, actions that exist in both $V_M$ and $V_E$ yield high similarity values in the corresponding regions of $S_{map}$. On the other hand, since the action of checking the connected equipment at the outlet box (prior to switching off the electrical panel breaker) does not exist in any part of $V_M$, the similarity observed in the corresponding column direction of $S_{map}$ is low. Thus, the result confirms that the similarity map $S_{map}$ generated by the proposed method is capable of representing the regions corresponding to candidate actions that incorporate the expert know-how from the work video created based on the manual and the work video of expert.

Table I shows the evaluation results on the 11 pairs of collected data obtained using both the proposed method and conventional methods. As shown in Table I, the proposed method is capable of extracting candidate actions that include expert know-how with an accuracy of 66.9%, which is 50 percentage points higher than the conventional methods. Furthermore, the number of frames that failed to be extracted is also smaller compared to the conventional methods. This indicates that the regions corresponding to candidate actions incorporating expert know-how extracted by the proposed method are more accurate than those extracted by conventional methods.

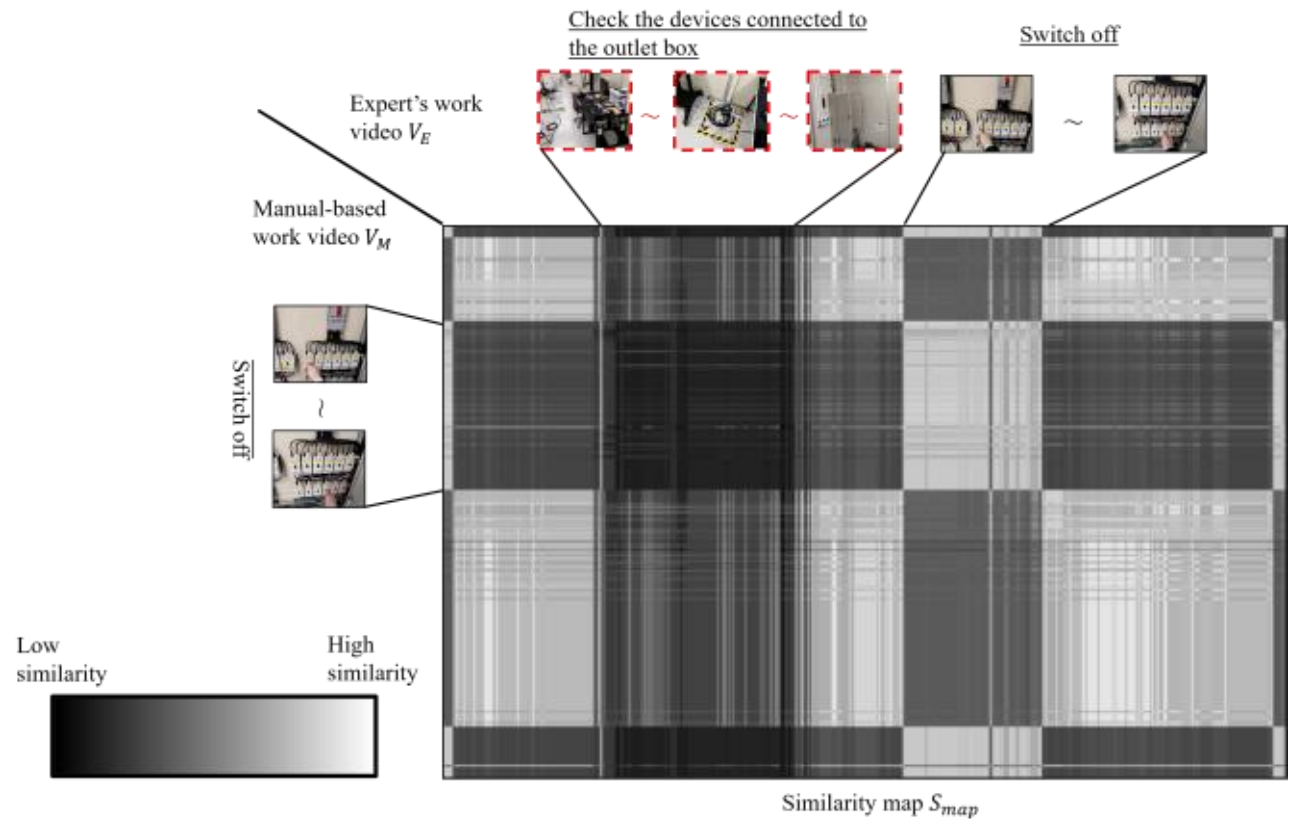

**FIGURE 6. Example of similarity map $S_{map}$ generated using the proposed method from $V_M$ and $V_E$.**

TABLE I
EVALUATION RESULTS ON COLLECTED DATA USING THE PROPOSED METHOD AND THE BASELINE METHODS

| Method | $Acc_{union}$ | Failure frame count |
|---|---|---|
| Proposed method | 66.92 % | 213.45 |
| Baseline (HLLM + Llama3.1) | 7.29 % | 443.73 |
| Baseline (HLLM + Llama3.2) | 9.33 % | 511.27 |
| Baseline (InternVideo2 + Llama3.1) | 16.12 % | 863.91 |
| Baseline (InternVideo2 + Llama3.2) | 13.81 % | 604.73 |

TABLE II
EVALUATION RESULTS FOR DIFFERENT VALUES OF THRESHOLD $\tau$

| Threshold $\tau$ | $Acc_{union}$ | Failure frame count |
|---|---|---|
| 0.0 | 0.00 % | 361.18 |
| 0.1 | 0.00 % | 361.18 |
| 0.2 | 0.00 % | 361.18 |
| 0.3 | 19.96 % | 262.91 |
| 0.4 | 38.52 % | 169.00 |
| 0.5 | 54.66 % | **121.36** |
| 0.6 | **66.92 %** | 213.45 |
| 0.7 | 55.42 % | 489.91 |
| 0.8 | 51.02 % | 611.55 |
| 0.9 | 42.92 % | 795.73 |
| 1.0 | 41.69 % | 875.09 |

The main difference between the conventional method based on HLLM and the proposed method lies in the approach used to extract using captions generated for the images. The proposed method vectorizes the captions, then creates a similarity map using the explicit metric of cosine similarity, and finally extracts the anomalous frames by employing the proposed filtering method. The conventional method, on the other hand, inputs the captions into a LLM and relies on the LLM to extract the anomalous frames. While the LLM makes a judgment as to whether a frame is anomalous based on the captions, the criteria for that judgment are not clearly defined, which is likely to result in lower extraction accuracy. In contrast, the proposed method clearly compares the captions through the calculation of similarity after vectorization, and the criteria for extracting anomalous frames can be determined based on the features of the entire video, as shown by Equation (3) in Algorithm 1. Therefore, it is considered that the proposed method is able to utilize the overall similarity between the two videos, rather than features based solely on individual images, to extract anomalous frames.

Regarding the conventional method, our results also confirm that while the extraction accuracy may be improved by using a more recent LLM, the improvement is small and there are cases where the accuracy decreases. In addition, the results from both the image-based and video-based conventional methods indicate that extracting anomalous frames based on the entire video rather than on individual images may improve the accuracy; however, the degree of improvement is also small.

Finally, as a detailed analysis of our method, we present the results of the sensitivity analysis for the threshold $\tau$ (Table II) to clarify the rationale for using $\tau = 0.6$. When $\tau$ is extremely small (0.0 to 0.2), the method is overly strict, failing to detect any anomalous frames and resulting in an $Acc_{union}$ of 0. As $\tau$ increases, the detection capability activates, and the performance significantly improves, peaking in the range of 0.5 to 0.6. Specifically, $\tau = 0.5$ yields the minimum number of failed frames (121.36), representing a conservative detection that effectively suppresses false positives. Conversely, $\tau = 0.6$ achieves the highest $Acc_{union}$ (66.9%), representing a more aggressive detection that broadly covers the ground truth intervals, albeit with an increase in over-detection. When $\tau \geq 0.7$, the method becomes overly sensitive to trivial localized variations, leading to excessive false positives and a drop in performance. For the practical application of extracting candidate know-how actions, missing an expert's unique operational step is highly undesirable. Therefore, prioritizing the comprehensive coverage of ground truth intervals over strict false-positive suppression, we adopted $\tau = 0.6$ for our main comparative evaluations. Importantly, the results demonstrate that our proposed method consistently outperforms the baseline methods (which achieved $Acc_{union}$ scores of 7.29% to 16.12%) across a wide operating window of $\tau \in [0.3, 0.8]$. This confirms that our global mean-based filtering is highly robust against localized variations and does not rely on overly strict parameter tuning.

## VII. DISCUSSION

In this study, we proposed a framework for extracting candidate actions that embody expert workers' tacit know-how by detecting anomalous frames between manual-based and expert work videos. By generating a similarity map using a pre-trained VLM, our method successfully identifies differences—such as independently added or appropriately skipped actions—without the need for task-specific training.

Regarding the method's behavior under temporal variations between videos, such as pacing differences or extra introductory frames, our approach computes the maximum similarity across the entire reference video for each frame (as shown in Algorithm 1). Therefore, it does not strictly assume frame-by-frame temporal alignment or identical video lengths. Consequently, simple variations in execution speed (pacing) or timing do not inherently mislead the similarity map, as the corresponding actions will still yield high maximum similarities regardless of their temporal location.

However, it should be noted that an expert's unique know-how might sometimes manifest precisely in these temporal variations—such as executing a task at a specific speed or timing. While our current method is designed to be robust against such variations to reliably detect added or omitted actions, capturing these subtle temporal differences as a form of know-how could be a valuable challenge for future work.

We also acknowledge that dealing with highly unconstrained real-world videos containing significant, non-informative temporal variations is a practical deployment challenge. To further enhance the robustness of our system in such scenarios, a possible future direction is to introduce action segmentation techniques on caption sequences prior to the similarity map generation, thereby filtering out irrelevant segments and focusing solely on task-related temporal boundaries.

Another important consideration for practical deployment is the computational cost. Generating a caption for every frame and constructing a full frame-by-frame similarity map results in high computational complexity, which can be challenging for high frame-rate or long videos. As the primary focus of this study was to realize zero-shot anomaly detection using a straightforward approach, optimizing computational efficiency was beyond our current scope. We acknowledge this computational burden as a limitation for large-scale real-world applications. To address this, frame sampling techniques could be effectively employed in future implementations to reduce the number of processed frames and save computation time. Furthermore, segmenting videos into appropriate temporal clip units prior to comparison is also considered a beneficial approach. Comparing videos at the level of meaningful temporal clips could simultaneously reduce computational cost and enhance the overall robustness of the system.

In this study, we employed BLIP-2 as a representative model due to its strong generalization capability. Although such models demonstrate strong performance, they may occasionally generate captions that are not fully consistent with the actual industrial environment. Furthermore, as demonstrated in the ablation study, the use of cleaned captions cleaned via Equation (2) improves the robustness of anomaly detection, indicating that the proposed framework is resilient to occasional captioning inconsistencies. Even if some inconsistencies persist after the correction step, they do not significantly affect the overall performance. This is because the proposed method relies on relative similarity patterns across the entire video rather than individual captions. As a result, sporadic captioning errors are less likely to distort the global structure of the similarity map.

Although the proposed framework was evaluated on only 11 video pairs, the method is zero-shot and training-free. Therefore, it does not rely on dataset-specific parameter fitting. Nevertheless, the scope of the current validation remains limited. The data are restricted to first-person maintenance videos recorded in relatively stable indoor environments, and each video pair contains at most one additional action or one omitted action. Extending the dataset to include more diverse industrial tasks and unseen environments remains an important direction for future work.

## VIII. CONCLUSION

In this study, we presented a novel method for automatically detecting anomalous frames within candidate actions, which may embody expert know-how in the field of infrastructure maintenance inspection. Our approach compares videos of tasks performed according to manuals with videos of expert operations to identify deviations that capture the unique tacit knowledge of experts. We achieve this by generating a similarity map between video frames using a pre-trained vision language model. The method then detects anomalous frames that likely represent expert-specific actions, either through the addition of necessary steps or the omission of redundant ones.

Our experiments on 11 types of actions not described in the manual in distribution board maintenance tasks demonstrate that the proposed method achieves a 66.9% detection rate. This performance marks an improvement of approximately 50 percentage points over a conventional baseline approach. These results suggest that our method is effective in codifying expert work know-how, with the potential to shorten the training period for new maintenance inspectors by providing explicit guidelines derived from expert practices.

Future work will focus on further enhancing detection accuracy, expanding the approach to include a broader range of tasks, and exploring integration with existing maintenance management systems. Advancing this study may contribute to the systematic education of maintenance procedures and help alleviate the shortage of expert maintenance workers.

## REFERENCES

[1] Recruit Works Institute, "Future Predictions 2040 in Japan," [Online]. Available: https://www.gii.co.jp/report/tbrc1465862-power-plant-maintenance-global-market-report.html. [Accessed 11 March 2025].

[2] E. Fenoglio, E. Kazim, H. Latapie and A. Koshiyama, "Tacit knowledge elicitation process for industry 4.0," Discover Artificial Intelligence, 2022.

[3] International Atomic Energy Agency, "Knowledge Management for Nuclear Industry Operating Organizations," International Nuclear Information System & Nuclear Knowledge Management Section International Atomic Energy Agency, 2006.

[4] M. Nakanishi , Y. Kido, “Externalization and Formalization of "Tacit Knowledge": Case of Skills Transfer of Pilot Boat Operations,” Sanno University Bulletin, 2017.

[5] S. Kernan Freire, C. Wang, S. Ruiz-Arenas and E. Niforatos, "Tacit Knowledge Elicitation for Shop-floor Workers with an Intelligent Assistant," CHI Conference on Human Factors in Computing Systems, 2023.

[6] R. Mancy and N. Reid, "Using interviews to investigate implicit knowledge in computer programming," Proceedings of the 7th international conference on Learning sciences, 2006.

[7] R. R. Hoffman, N. R. Shadbolt, A. Burton and G. Klein, "Eliciting knowledge from experts: A methodological analysis," Organizational behavior and human decision processes, 1995.

[8] J. Summerscales, "Harvesting tacit knowledge for composites workforce development," Composites Part A: Applied Science and Manufacturing, 2024.

[9] T. Gavrilova, "Knowledge elicitation techniques in a knowledge management context," Journal of Knowledge Management, 2012.

[10] R. R. Hoffman, "A survey of methods for eliciting the knowledge of experts," ACM SIGART Bulletin, 1989.

[11] N. Shadbolt and P. R. Smart, "Knowledge Elicitation: Methods, Tools and Techniques," 2015. [Online]. Available: https://api.semanticscholar.org/CorpusID:18724737.

[12] T. Nagarajan and L. Torresani, "Step Differences in Instructional Video," Proceedings of the IEEE/CVF Conference on Computer Vision and Pattern Recognition, 2024.

[13] C. Jiang, K. Huang, S. He, X. Yang, W. Zhang, X. Zhang, Y. Cheng, L. Yang, Q. Wang, F. Xu, T. Pan and W. Chu, "Learning Segment Similarity and Alignment in Large-Scale Content Based Video Retrieval," Proceedings of the 29th ACM International Conference on Multimedia, 2021.

[14] G. Kordopatis-Zilos, G. Tolias, C. Tzelepis, I. Kompatsiaris, I. Patras and S. Papadopoulos, "Self-Supervised Video Similarity Learning," Proceedings of the IEEE/CVF Conference on Computer Vision and Pattern Recognition Workshops, 2023.

[15] S. He, Y. He, M. Lu, C. Jiang, X. Yang, F. Qian, X. Zhang, L. Yang and J. Zhang, "TransVCL: Attention-enhanced Video Copy

Localization Network with Flexible Supervision," 37th AAAI Conference on Artificial Intelligence, 2023.
[16] Z. Han, X. He, M. Tang and Y. Lv, "Video Similarity and Alignment Learning on Partial Video Copy Detection," Proceedings of the 29th ACM International Conference on Multimedia, 2021.
[17] T. Yunlong, B. Jing, X. Siting, S. Luchuan, L. Susan, W. Teng, Z. Daoan, A. Jie, L. Jingyang, Z. Rongyi, V. Ali, H. Chao, Z. Zeliang, L. Pinxin, F. Mingqian, Z. Feng, Z. Jianguo, L. Ping, L. Jiebo and X. Chenliang, "Video Understanding with Large Language Models: A Survey," IEEE Transactions on Circuits and Systems for Video Technology, 2025.
[18] W. Mengmeng, X. Jiazheng and L. Yong, "ActionCLIP: A New Paradigm for Video Action Recognition," arXiv, 2021.
[19] C. Guo, Z. Yin-Dong, W. Jiahao, X. Jilan, H. Yifei, P. Junting, W. Yi, W. Yali, Q. Yu, L. Tong and W. Limin, "VideoLLM: Modeling Video Sequence with Large Language Models," arXiv, 2023.
[20] L. Zanella, W. Menapace, M. Mancini, Y. Wang and E. Ricci, "Harnessing Large Language Models for Training-free Video Anomaly Detection," Proceedings of the IEEE/CVF Conference on Computer Vision and Pattern Recognition, 2024.
[21] Trimble Inc., "Trimble XR10 with HoloLens 2," 2022. [Online]. Available: https://fieldtech.trimble.com/en/products/mixed-reality/trimble-xr10-with-hololens-2.
[22] D. Ungureanu, F. Bogo, S. Galliani, P. Sama, X. Dua, C. Meekhof, J. Stühmer, T. J.Cashman, B. Tekin, J. L. Schönberger, P. Olszta , M. Pollefeys, "HoloLens 2 Research Mode as a Tool for Computer Vision Research," arXiv, 2020.
[23] J. Li, D. Li, S. Savarese and S. Hoi, "BLIP-2: bootstrapping language-image pre-training with frozen image encoders and large language models," Proceedings of the 40th International Conference on Machine Learning, 2023.
[24] G. Rohit, E.-N. Alaaeldin, L. Zhuang, S. Mannat, A. Kalyan Vasudev, J. Armand and M. Ishan, "ImageBind: One Embedding Space To Bind Them All," Proceedings of the IEEE/CVF Conference on Computer Vision and Pattern Recognition, 2023.
[25] Llama team, "The Llama 3 Herd of Models," arXiv, 2024.
[26] Meta, "lama-3.1-8B-Instruct," 23 July 2024. [Online]. Available: https://huggingface.co/meta-llama/Llama-3.1-8B-Instruct. [Accessed 6 March 2025].
[27] Meta, "Llama-3.2-3B-Instruct," 25 September 2024. [Online]. Available: https://huggingface.co/meta-llama/Llama-3.2-3B-Instruct. [Accessed 6 March 2025].
[28] Y. Wang, K. Li, X. Li, J. Yu, Y. He, C. Wang, G. Chen, B. Pei, Z. Yan, R. Zheng, J. Xu, Z. Wang, Y. Shi, T. Jiang, S. Li, H. Zhang, Y. Hunag, Y. Qiao, Y. Wang and L. Wang, "InternVideo2: Scaling Foundation Models for Multimodal Video Understanding," European Conference on Computer Vision, 2024.

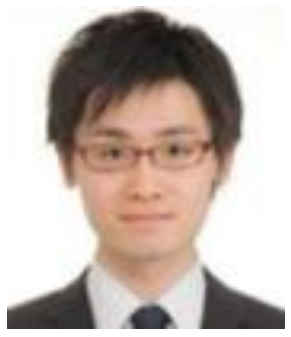

**RYO SAKAI** received Master of Information Science from Osaka University in 2016. He is currently a senior researcher at the Robotics Research Department in Research and Development Group, Hitachi, Ltd. His research interest includes computer vision, machine learning, tacit knowledge extraction, robotics, and intelligent system. He is a member of the Robotics Society of Japan.

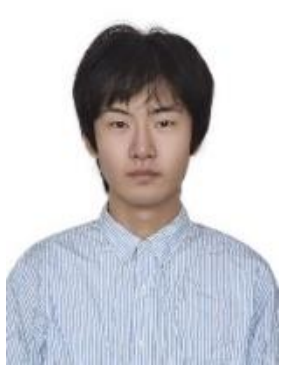

**YONGPENG CAO** is a Ph.D student at the University of Tokyo. He received a Master degree of Mechanical Engineering in the University of Tokyo in 2022 a Bachelor degree in Mechanical Engineering from Beijing University of Chemical Technology in 2020. His research interests include human-robot Interaction, human action recognition and high-speed vision system.

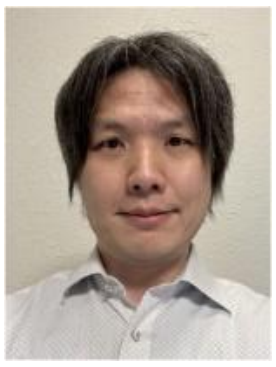

**NOBUTAKA KIMURA** is a manager at the Next Research Department in Research & Development Group of Hitachi, Ltd. He received a Bachelor degree and a Master degree in mechanical science and engineering from Tokyo Institute of Technology in 2003 and 2005, respectively. In 2005, he joined the Advanced Research Laboratory at Hitachi, Ltd. After joining the company, he also received a PhD in Engineering from the University of Tokyo in 2017. From 2022 to 2023, he was temporarily transferred to Hitachi America, Ltd. His research interests include robotic vision and system integration for autonomous robotic systems in warehouses, factories and retail stores.